\documentclass[conference]{IEEEtran}
\usepackage{amsmath}
\usepackage{mathtools}
\usepackage{cite}
\usepackage{cases}
\usepackage{graphics}
\usepackage{array}
\usepackage{float}
\usepackage{epstopdf}
\usepackage{tabularx}
\usepackage{subfigure}
\usepackage{listings}
\usepackage{booktabs}
\usepackage{tabularx}
\usepackage{float}
\usepackage{threeparttable}
\usepackage{flushend,cuted}

\ifCLASSINFOpdf
\else
\fi
\begin{document}
%

\title{Sparse Deep Nonnegative Matrix Factorization}



%
\author{\IEEEauthorblockN{Zhenxing Guo\IEEEauthorrefmark{1} and
Shihua Zhang\IEEEauthorrefmark{1}\IEEEauthorrefmark{2}\IEEEauthorrefmark{3}
}
\IEEEauthorblockA{\IEEEauthorrefmark{1}National Center for Mathematics and Interdisciplinary Sciences\\
Academy of Mathematics and Systems Science, Chinese Academy of Sciences, Beijing 100190, China\\
}
\IEEEauthorblockA{\IEEEauthorrefmark{2}School of Mathematics Sciences, University of Chinese Academy of Sciences, Beijing 100049, China\\
}
\IEEEauthorblockA{\IEEEauthorrefmark{3}To whom correspondence should be addressed. Email: zsh@amss.ac.cn}
}



\maketitle


\begin{abstract}
Nonnegative matrix factorization is a powerful technique to realize dimension reduction and pattern recognition through single-layer data representation learning. Deep learning, however, with its carefully designed hierarchical structure, is able to combine hidden features to form more representative features for pattern recognition. In this paper, we proposed sparse deep nonnegative matrix factorization models to analyze complex data for more accurate classification and better feature interpretation. Such models are designed to learn localized features or generate more discriminative representations for samples in distinct classes by imposing $L_1$-norm penalty on the columns of certain factors. By extending one-layer model into multi-layer one with sparsity, we provided a hierarchical way to analyze big data and extract hidden features intuitively due to nonnegativity. We adopted the Nesterov's accelerated gradient algorithm to accelerate the computing process with the convergence rate of $O(1/k^2)$ after $k$ steps iteration. We also analyzed the computing complexity of our framework to demonstrate their efficiency. To improve the performance of dealing with linearly inseparable data, we also considered to incorporate popular nonlinear functions into this framework and explored their performance. We applied our models onto two benchmarking image datasets, demonstrating our models can achieve competitive or better classification performance and produce intuitive interpretations compared with the typical NMF and competing multi-layer models.
\end{abstract}


%
\IEEEpeerreviewmaketitle

\section{Introduction}
\noindent
Nonnegative matrix factorization (NMF)is a powerful dimension reduction and pattern recognition technique in data analysis \cite{Lee1999}, which has been widely used in diverse areas such as document clustering \cite{Xu2003},\cite{Pauca2004},\cite{Shahnaz2006}, face recognition \cite{Wang2005},\cite{Guillamet2002} and microarray data analysis \cite{Brunet2004},\cite{Qi2009}. It decomposes a nonnegative matrix $X$ into the product of two low-rank nonnegative factor matrices $W$ and $H$ (i.e., $X\approx WH$ with $W\geq0$, $H\geq0$), generating natural interpretations in many cases. Moreover, NMF has been extended into a number of variant forms, allowing for various sparse \cite{Hoyer2004},\cite{Kim2007},\cite{Peharz2012} or regularized models \cite{Cai2008},\cite{Cai2011}, most of which demonstrate distinct advantages in local feature extraction or data representation learning.

For sparse variants of NMF, Hoyer \cite{Hoyer2004} found that the original NMF does not always obtain part-based features. Thus, they proposed new sparse models to explicitly control the degree of sparseness of $W$ or $H$, and the new models can learn part-based features that cannot be revealed by the original NMF. However, Kim and Park \cite{Kim2007} proposed that if strong sparsity constraints are imposed, some important information for gene selection in microarray analysis may be lost. They therefore established a varied version of sparse NMF model, in which, $L_1$-norm penalty was adopted to constrain the sparseness of columns of $W$ or $H$ and this constriction was added as one term to the objective function. This model has been demonstrated to be able to realize biclustering for cancer subtype discovery and gene expression analysis. Peharz and Pernkopf \cite{Peharz2012} presented another sparse NMF version by adopting $L_0$-norm penalty to limit the exact number of zeros in $W$ or $H$. Besides, graph-regularized NMF versions have also been explored. For example, Cai \textit{et al.} \cite{Cai2011} proposed a graph-regularized NMF by incorporating prior information of samples into the typical NMF. This helps to keep the original topological structure of data after being projected into a subspace and usually leads to better clustering results.

Moreover, NMF has also been extended for data fusion and combinatorial patterns extraction when analyzing multiple data. For example, Zhang \textit{et al.} \cite{Zhang2012} developed a joint non-negative matrix factorization (jNMF) technique to integrate multi-dimensional genomics data for the discovery of combinatorial patterns to reveal phenomena that would have been ignored with only a single type of data. This model provides a way to reveal the homogeneous relationships among different data. Furthermore, Zhang \textit{et al.} extended the jNMF to both sparse and graph-regularized version \cite{Zhang2011}. Unlike jNMF, Yang and Michailidis \cite{Yang2016} proposed integrative NMF (iNMF) model, besides the homogeneous effects, they also considered the heterogeneous effects of different type of data. Lastly, $\check{Z}$itnik and Zupan \cite{Zitnik2015} proposed a matrix factorization-based data fusion method DFMF to integrate relationships of heterogeneous datasets and describe the observed system from various views to reveal hidden associations.

Although there have been extensive variants of NMF, most of them remain to be single-layer models. Deep learning is becoming increasingly popular and has been demonstrated to be powerful in learning data representation \cite{Hinton2006_1}, \cite{Hinton2006_2}, \cite{Mohamed2009}, \cite{Hinton2012_1}, \cite{Hinton2012_2}. However, typical deep learning is rather complicated, which requires complex structures, needs empirical skills to tune a lot of parameters and lacks explicitly theoretical foundations. Recently, several new frameworks (e.g., Deep PCA (DPCA) \cite{Liong2013}, PCANet \cite{Chan2015} and gcForest \cite{Zhou2017}) have been proposed to attempt to tackle these issues and provide alternative views to deep learning. DPCA \cite{Liong2013} performs a two-layer Zero-phase Component Analysis (ZCA) whitening plus PCA process to learn hierarchical features and obtain corresponding representations for face recognition. PCANet \cite{Chan2015} contains a two-layer architecture, where PCA is employed in each layer to learn multistage filterbanks. The deep architecture is followed by binary hashing and block histograms for indexing and pooling to generate deep features for more accurate classification. gcForest \cite{Zhou2017} assembles different type of random forests at each layer to learn deep features to realize comparable classification performance with deep neural network or convolutional neural network, without establishing complex structures and tuning extensive parameters like traditional deep learning. Inspired by the success of them, multi-layer factorizations are attractive to break down the complex problem hierarchically into multiple simple ones. Along these lines, Multi-layer NMF \cite{Ahn2004}, \cite{Song2015} and Deep Semi-NMF \cite{Trigeorgis2017} have been proposed recently. The general idea of them is by stacking one-layer NMF or Semi-NMF \cite{Ding2010} into multiple layers to learn hierarchical relationships among features or hierarchical projections. However, these methods do not well reflect \cite{Song2015} or even ignore the sparse structure hidden in the complex data \cite{Ahn2004}, \cite{Trigeorgis2017}.

To this end, we developed sparse deep NMF models and explored their effectiveness in multiple aspects. In the new models, the first layer is responsible for breaking down the original data into multiple initial basis. Then the factorization in the second layer is to generate meaningful relationships among all the basis in the first layer to form relatively complex features. Again, the decomposition in the third or higher layer is to learn relationships among features from the former layer to combine them selectively. This process is repeated until to the highest layer. In the end, besides all relationship matrices, a data representation matrix will be automatically yielded. During the factorization in each layer, different sparsity constraints were added to localize partial features or generate more discriminative representations for samples in different classes. The key idea is that by thoroughly learning the hidden basis as well as the additional relationships across layers, we can obtain a high level data representation matrix and yield intuitive interpretations for features generated in each layer.

We summarized the main contributions of this paper as follows:
\begin{itemize}
  \item We proposed a series of generalized sparse deep NMF models by extending single-layer algorithm into multiple layers, each of which is designed to generate matrices satisfying certain sparsity requirements.
  \item We also considered linearly inseparable data by incorporating nonlinear functions into the deep NMF models with different ways.
  \item We adopted the Nesterov's accelerated gradient descent algorithm \cite{Nesterov2005} to accelerate the optimization process with convergence rate $O(\frac{1}{k^{2}})$, which is much faster than traditional gradient descend algorithm with convergence rate $O(\frac{1}{k})$.
\end{itemize}

The organization of this paper is as follows. In section 2, we introduced highly related works and explained their characteristics. In section 3, we proposed a series of diverse models as well as their corresponding optimization algorithms. Besides, we explored the effectiveness of incorporation of different nonlinear functions into these models. In section 4, we employed two benchmarking datasets, ORL and PIE-pose 27.0, to demonstrate the properties and performance of our models. We showed that the ORL data (consisting of only 400 images) with a relative small number of samples are not sufficient to train a deep model. With the PIE-pose 27.0 data consisting of 2856 images, we conducted extensive experiments and compared them with other NMF variants to demonstrate the effectiveness of our models. We also tested how the structure of deep models, the number of layers and the number of subbasis in hidden layers affect the performance of our models. Finally, we concluded this paper in section 5.

\section{Related work}
\noindent
In this section, we introduce three related studies including two deep frameworks of matrix factorization models and the Nesterov's accelerated gradient descent algorithm for solving the typical NMF. Specially,
\subsection{Multi-layer NMF}
\noindent
The Multi-layer NMF model \cite{Song2015} extends the typical NMF explicitly to multiple layers in the following format:
\begin{eqnarray}\label{Multi-layer}
                      &  \min     &\frac{1}{2}\parallel X-W_1W_2W_3...W_LH_L\parallel^2_F \notag \\
                                            \\
                   & \mbox{s.t.}  & \widetilde{H}_{l-1}\approx W_ l \widetilde{H}_l, \notag \\
                   &           &   W_l \geq 0,  \widetilde{H}_l \geq 0, \quad l=1,2,...,L. \notag
\end{eqnarray}
where $L$ is the number of layers. When $L=1$, it reduces to the typical NMF model. Ahn \textit{et al.} \cite{Ahn2004} first proposed a multi-layer NMF model coupled with three nonlinear layers. They developed a strategy called up-propagation algorithm to jointly update all weight matrices and demonstrated its role in extracting hierarchical features. Song \textit{et al.} \cite{Song2015} proposed one multi-layer NMF model and adopted non-smooth NMF (nsNMF) \cite{Pascual2006} to solve the typical NMF in each layer. nsNMF utilizes one sparse matrix $S$ to apply sparsity constraint to standard NMF: $S=(1-\theta)I(k)$+$\frac{\theta}{k} $ones$(k)$. $k$ is the number of bases in corresponding layer, and $\theta$ is parameter for smoothing effect, in the range of 0 to 1. $I(k)$ is identity matrix of size $k\times k$ with all components of 1s. nsNMF smoothes a matrix by multiplying it with $S$. In Multi-layer NMF, the author smoothed $H$ matrix by multiplying $S$ and $H$ during iterations as $H=SH$. Then $W$ would be sparse to compensate the loss of sparsity. They applied their model to the Reuters-21578 collection dataset and showed its superiority in classification and feature hierarchies interpretations compared with the single-layer nsNMF.

\subsection{Deep Semi-NMF}
\noindent
Compared to the typical NMF, Semi-NMF does not require the original data and the basis matrix to be nonnegative. It extends the applicable range of NMF, but generates basis matrices that hardly show parts-of-whole characteristics intuitively\cite{Ding2010}. Deep Semi-NMF \cite{Trigeorgis2017} is an extension of Semi-NMF by considering the matrix factorization in a multi-layer manner. It is defined in the following format,
\begin{eqnarray}\label{Deep Semi-NMF}
                      &  \min     &\frac{1}{2}\parallel X^{\pm}-W^{\pm}_1W^{\pm}_2...W^{\pm}_LH^{+}_L\parallel^2_F \notag \\
                                                                                         \\
                      &  \mbox{s.t.}  & H^{+}_{l-1}\approx W^{\pm}_ l H^{+}_l,   \notag \\
                      &           & H_l \geq 0, \quad l=1,2,...,L. \notag
\end{eqnarray}
This method first pre-trains each layer, then fine-tunes the whole system with the outcomes of the pre-training. In general, without nonnegative constraint, $W^{\pm}_l$ can be directly updated by the least square algorithm. Nonnegative $H^{+}_l$ is updated in a multiplicative manner with the help of auxiliary function like that in \cite{Ding2010}. Deep Semi-NMF turns out to be useful in learning hierarchical projections, from the original data space to various subspaces spanned by hidden attributes (e.g., face expression, illumination angle for face image data). Thus, it can cluster samples according to their hidden attributes in corresponding layers.

Deep Semi-NMF does not require the input data and the basis matrices to be nonnegative. Thus, it is hard to see the part-of-whole phenomenon because mutually offset occurs when the basis matrices are combined. Sparsity constraints were not considered either. Besides, the learning process that extracts the hierarchical projections from raw data space to multiple hidden attributes subspace automatically is too abstract to understand. For face images, attributes like face expression, photographing angle etc are easy to be observed but the hidden attributes of many other kinds of data are hardly to know.

On the other hand, Ahn \textit{et al.} \cite{Ahn2004} described an interesting multiple decomposition for nonnegative matrix, but neglected detailed analysis for its structure. Besides, sparsity structure was not considered. As for the algorithm, they updated the feature matrices jointly without layer-wise initialization, so the results may vary greatly from one to another experiment. Song \textit{et al.} \cite{Song2015} chose the proper number of initial basis vectors through single-layer nsNMF. They adopted nsNMF and attempted to control the sparsity level of one overall matrix by tuning parameter $\theta$, which cannot generate a column-wise sparse structure to localize features for each sample. Moreover, in their experiments, results obtained by $\theta=0$ are better than that obtained by $\theta\neq0$, implying that the sparsity in their research does not work. For the optimization strategies, they adopted traditional multiplicative update algorithm for nsNMF of each layer and then stacked it into two layers. They did not fine-tune the whole system to reduce the total reconstruction error. This workflow is simple but may suffer from the problem of low convergence. Moreover, it is sensitive to initial solutions, making it rather unstable.

However, by restricting the sparsity of each column of certain matrices, sparse structure in the original data is well explored in our sparse deep NMF models. Specifically, we considered the sparsity problem of all basis matrices $W_l$ to extract local features. We also tried to solve a sparse coding problem by constricting the sparsity of representation matrices $H_l$, representing the original data with as fewer basis vectors as possible while maintaining the discriminative ability of representations for samples in distinct classes. As for the optimization strategy, initialization is first proceeded for each layer, at the beginning of which, NNSVD \cite{Boutsidis2008} was adopted to generate a good pair of initialization factors. Then we fine-tuned the whole system to reduce the total reconstruction error. Nesterov's accelerated gradient descent algorithm \cite{Nesterov2005} was adopted to alleviate the problem of numerical instability and low convergence rate. We conducted a lot of numerical tests to optimize our model's structure. Popular nonlinear functions as well as the way to be incorporated was explored by extensive experiments.

\subsection{NeNMF}
\noindent
Nesterov's accelerated gradient descent algorithm \cite{Nesterov2005} was adopted by Guan \textit{et al.} to develop an efficient NMF solver NeNMF \cite{Guan2012}. It has been demonstrated to be able to alleviate the problem of numerical instability and low convergence rate, frequently accounted by other NMF algorithms. Here, we gave a brief introduction to NeNMF. The spirit of NeNMF was also adopted in our algorithms. It is known that NMF is a nonconvex optimization problem and it is impractical to obtain the optimal solution. Thus, block coordinate descent method \cite{Bertsekas1999} is popular to seek for a local optimum. Given an initial pair of $W^1$ and $H^1$, the block coordinate descent method alternatively solves
\begin{eqnarray}\label{NeNMF_H}
                       H^{t+1} = &\arg \min_{H\geq0}F(W^t,H) = \frac{1}{2}\parallel X-W^t H\parallel^2_F,
\end{eqnarray}
and
\begin{eqnarray}\label{NeNMF_W}
                       W^{t+1} = &\arg \min_{W\geq0}F(W,H^t) = \frac{1}{2}\parallel X-W H^t\parallel^2_F,
\end{eqnarray}
until convergence, where $t$ is the iteration step. NeNMF employs Nesterov's accelerated gradient descent algorithm to solve \eqref{NeNMF_H} and \eqref{NeNMF_W}. Take $H$ for instance, at each iteration, $H$ is updated by the projected gradient method, and it performed on a chosen search point. The step size is determined by the Lipschitz constant not by the time consuming line search. Due to the convexity of \eqref{NeNMF_H} with respect to $H$ and the continuity of the corresponding gradient, the convergence rate is $O(\frac{1}{k^2})$ after $k$ steps in iteration, superior to conventional gradient descend algorithm whose convergence rate is $O(\frac{1}{k})$.

In particular, two sequences (i.e., ${H_k}$ and $Y_k$) are constructed and updated alternatively in the following way,
\begin{eqnarray}\label{NeNMF_H approximal fun}
                     & H^{t+1} = &\arg \min_{H\geq0} \{\Phi(Y_k,H)=F(W^t,Y_k) \\
                     &           & +\langle \nabla_H F(W^t,Y_k),H-Y_k \rangle + \frac{LC}{2}\parallel H-Y_k\parallel_F^2\}, \notag
\end{eqnarray}
\begin{eqnarray}\label{NeNMF_Y update formula}
                     Y_{k+1}=H_k+\frac{\alpha_k-1}{\alpha_{k+1}}(H_k-H_{k-1}),
\end{eqnarray}
where $\Phi(Y_k,H)$ is the approximate function of $F(W^t,H)$ on $Y_k$, $LC$ is the corresponding Lipschitz constant. $H_k$ contains the approximate solution obtained by minimizing $\Phi(Y_k,H)$ over $H$, and $Y_k$ stores the search point that is constructed by linearly combining the latest two approximate solutions, i.e., $H_{k-1}$ and $H_k$. According to \cite{Nesterov2005}, the combination coefficient is carefully updated in each iteration step as follows,
\begin{eqnarray}\label{NeNMF_alpha}
                    \alpha_{k+1}=\frac{1+\sqrt{4\alpha_k^2+1}}{2}.
\end{eqnarray}
Based on the Lagrange multiplier method, the Karush-Kuhn-Tucker (K.K.T.) conditions of \eqref{NeNMF_H approximal fun} are as follows,
\begin{eqnarray}\label{NeNMF H KKT}
                    \nabla_H\phi(Y_k,H_k) & \geq 0, \notag \\
                                      H_k & \geq 0, \\
        \nabla_H\phi(Y_k,H_k) \otimes H_k & \geq 0, \notag
\end{eqnarray}
where $ \nabla_H\phi(Y_k,H_k) = \nabla_H\phi(W^t,Y_k)+ LC(H_k-Y_k)$ is the gradient of $\phi(Y_k,H)$ with respect to $H$ at $H_k$, and $\otimes$ is the Hadamard product. By solving \eqref{NeNMF H KKT}, we can obtain the update formula
\begin{eqnarray}\label{NeNMF H_update formula}
                      H_k = P \left(Y_k-\frac{1}{LC}\nabla_H F(W^t,Y_k)\right),
\end{eqnarray}
where the operator $P(X)$ projects all the negative entries to zeros. By alternatively updating $H_k$, $Y_{k+1}$ and $\alpha_{k+1}$ with \eqref{NeNMF H_update formula}, \eqref{NeNMF_Y update formula} and \eqref{NeNMF_alpha} respectively until convergence, the optimal solution of \eqref{NeNMF_H} can be reached. As \eqref{NeNMF_W} and \eqref{NeNMF_H} are symmetrical, the update for $W^{t+1}$ will be obtained in a similar way. NeNMF converges fast eventually with above optimization strategy.

\section{Sparse Deep NMF}
\noindent
Similar to the general multi-layer NMF framework, sparse deep NMF models factorize a nonnegative matrix into $L+1$ nonnegative ones:
$$X \approx W_1g^{-1}(W_2\cdot\cdot\cdot g^{-1}(W_LH_L )).$$
To make it more intuitive, we can split the equation into the following formula:
                        \[ g(H_{L-1}) \approx W_ L H_L, \]
                        \[ g(H_{L-2}) \approx W_ {L-1} g^{-1}(W_ L H_L),  \]
                        \[......,\]
                        \[ g(H_2) \approx W_3g^{-1}( \cdot\cdot\cdot g^{-1}( W_L H_L)), \]
                        \[ g(H_1) \approx W_2 g^{-1}(\cdot\cdot\cdot g^{-1}( W_L H_L)), \]
                                                \[ X \approx W_ 1 H_1, \]
where $g$ can be a linear or nonlinear function if necessary. All of the matrices above are required to be nonnegative. For the decomposition in the first layer, $W_1$ is responsible for storing initial sub-basis matrix, the columns of which should be sufficient enough to learn as much information as possible. Then the further factorization on $H_1$ is to learn relationships among various sub-basis vectors in $W_1$ to combine them and form meaningful and decipherable features. Further decomposition on $H_l$ ($l=2,3,...,L-1$) serves as the similar function with the decomposition of $H_1$. With this hierarchical learning structure, a sequence of sub-basis matrices $W_l$ ($l=1,2,...,L$) are generated. Complex features of raw data are extracted by additionally combining those sub-basis matrices. Meanwhile, once the complex features are learned, a high level data representation will be more representative for samples, leading to more accurate classification.

In our sparse deep NMF models, we thoroughly considered the sparse structure hidden in complex data. We first imposed $L_1$-norm penalty onto each column of $W$ in each layer (denoted as SDNMF/L). We also considered to impose $L_1$-norm penalty onto each column of $H$ (denoted as SDNMF/R), which helps to solve sparse coding problem by approximating raw vector with as fewer bases as possible. SDNMF/L only imposes $L_1$-norm penalty on each $W_l$ ($l=1,2,...,L$) whereas SDNMF/R only imposes the sparsity of $H_l$. Next, we imposed sparse constraints on both $W_l$ and $H_l$. One model from this thought requires all factors to be sparse to generate sparse sub-bases and representations (denoted as SDNMF/RL1). Intuitively, for a decomposition of $X \approx WH$, if the $W$ is compelled to be sparse, $H$ tends to be smooth. We strengthened this tendency by controlling the $L_1$-norm of each column of $W_l$ while imposing $L_2$-norm constraints onto final $H_L$ (denoted as SDNMF/RL2). Also, we inspected the performance of model DNMF. It is one case of our models where there isn't any sparsity constraint on any factors.

\subsubsection{SDNMF/L}
Specifically, SDNMF/L is formulated as follows:
\begin{eqnarray}\label{SDNMF/L0}
                      &  \min     &\frac{1}{2}\parallel X-W_1g^{-1}(W_2...g^{-1}(W_LH_L))\parallel^2_F \notag\\
                      &           & +\frac{1}{2}\Sigma^L_{l=1}\mu_l\Sigma^n_{j=1}\parallel W_l(:,j)\parallel^2_1  \\
                                                                                                                     \notag \\
                 &  \mbox{s.t.}  & g(H_{l-1})\approx W_ l H_l,  \notag \\
                 &           &   W_l \geq 0, \quad H_l \geq 0, \quad l=1,2,...,L, \notag
\end{eqnarray}
where $\parallel W_l(:,j)\parallel_1 $ denotes $\Sigma_i \mid W_l(i,j)\mid $. Here, we adopted the square of $\parallel W_l(:,j)\parallel $ in case that a severe penalty would cause the loss of useful information. Assume that $x$ is one column vector of input data $X$; $h$ is the corresponding coefficient vector in $H_1$. For $W_1$, taking it as a linkage between the input vector $x$ and the corresponding representation vector $h$ in the first layer, the more sparse one of its column $w_k$ is, the fewer links there will be between $h_k$ (the $k$-th element in $h$) and $x$. This means that each element in $h$ will respond selectively to all the elements in $x$, or each of them will be only activated by the certain members of $x$, resulting partial information of $x$ being captured by $w_k$. This is consistent with the desire of extracting localized features for each sample. The sparsity constraints for $W_2$ aims at capturing local information in $H_1$ containing the rough relationships among all $w_k$ in $W_1$, helping to selectively combine certain $w_k$ to form meaningful and distinguishable features. The sparsity of $W$ in upper layer functions similarly with $W_2$.

Here we took $g(x)=x$ to illustrate the algorithm simply and nonlinear models later. It first pre-trains the model in a layer-wise way. For the $\emph{l}$-th layer, it optimizes the following model:
\begin{eqnarray}\label{sub_l}
                     & \min &\frac{1}{2}\parallel H_{l-1}-W_l H_l\parallel^2_F + \frac{1}{2}\mu_l \mbox{tr}((\xi_lW_l)^T(\xi_lW_l)) \notag \\
                     \\
                     & \mbox{s.t.} &W_l \geq 0, H_l \geq 0, \notag
\end{eqnarray}
where $H_0=X$, $\xi_l$ is a row vector with all components equal one. $\mbox{Tr}((\xi_lW_l)^T(\xi_lW_l))$ is a variant of $\parallel W_l(:,j)\parallel^2_1 $ due to the nonnegativity of $W$. Different optimization strategy can be used to optimize it. For example, projected gradient (PG) method \cite{Lin2007} takes advantage of the Armijo line search to estimate the optimal step size along the projection direction for solving each subproblem. However, the Armijo rule is a time-consuming search strategy, making PG inefficient. We can also consider the projected NLS method (PNLS) proposed by Berry \textit{et al.} \cite{Berry2007}. However, it is an approximate approach and cannot guarantee the convergence.

\begin{table}[htbp] \small
\caption{}\label{INI_SDNMF/L}
\centering  
\begin{tabular*}{250pt}{lccc}  
\hline
 Algorithm 1: Optimal gradient method for initialization \\ \hline
\textbf{Input}:        $H_{l-1}$\\
\textbf{Output}:       $ W_l,H_l$ \\
1. Initialize $ H_l^1\geq0$ and $W_l^1\geq0$ with NNSVD, $t=1 $\\
\textbf{Repeat}\\

2. Update $H_l^{t+1}$ and $W_l^{t+1}$ with \\
  \hspace{0.2cm}    2.1 $ H_l^{t+1}=\mbox{Subalgorithm 1}(W^t,H^t,\nabla_H F(W^t,H),LC_{H_l^t})$\\
  \hspace{0.2cm}    2.2 $ W_l^{t+1}=\mbox{Subalgorithm 2}(W^t,H^{t+1},\nabla_W F(W,H^{t+1}),LC_{W_l^t})$\\
3.    $ t\leftarrow t+1 $ \\

\textbf{Until} stopping criterion is satisfied\\
4.    $ W_l= W^t ,H_l=H^t $\\ \hline
\\
\hline
 Subalgorithm 1: Optimal gradient method for $H_l$ \\ \hline
\textbf{Input}:        $ W_l^t$, $H_l^t$\\
\textbf{Output}:       $ H_l^{t+1}$ \\
1. Initialize $H_0=H^t_l$, $Y_0=H^t_l$, $\alpha_0=1$, \\
\quad\quad $LC_{H_l^t}=\parallel (W_l^t)^T W_l^t\parallel_2+\lambda_l \parallel e_l^T e_l\parallel_2$, $k=0$\\
\textbf{Repeat}\\

2. Update $H_k$, $\alpha_{k+1}$ and $Y_{k+1}$ with\\
  \hspace{0.2cm}    2.1 $ H_k=P(Y_k-\frac{1}{LC_{H_l^t}}\nabla_H F(W_l^t, Y_k))$\\
  \hspace{0.2cm}    2.2 $ \alpha_{k+1}=\frac{1+\sqrt{4\alpha_k^2+1}}{2}$\\
  \hspace{0.2cm}    2.3 $Y_{k+1}= H_k + \frac{\alpha_k-1}{\alpha_{k+1}}(H_k-H_{k-1})$\\
3.    $ k\leftarrow k+1 $ \\

\textbf{Until} stopping criterion is satisfied\\
4.    $ H_l^{t+1}= H_k $\\ \hline
\\
\hline
 Subalgorithm 2: Optimal gradient method for $W_l$ \\ \hline
\textbf{Input}:        $ W_l^t$, $H_l^t$ ($\Psi_{l-1}^t$ need to be input when fine-tuning the system)\\
\textbf{Output}:       $ W_l^{t+1}$ \\
1. Initialize $ W_0=W^t_l$, $Z_0=W^t_l$, $\beta_0=1$, \\
 \quad\quad $ L_{W_l^t}=\parallel (H_l^t)(H_l^t)^T\parallel_2 + \mu_l \parallel \xi_l^T\xi_l\parallel_2$, $k=0$\\
\textbf{Repeat}\\

2. Update $W_k, \beta_{k+1}$ and $Z_{k+1}$ with\\
  \hspace{0.2cm}  2.1 $ W_k=P(Z_k-\frac{1}{LC_{W_l^t}}\nabla_W F(Z_k, H_l^t))$\\

  \hspace{0.2cm}  2.2 $ \beta_{k+1}=\frac{1+\sqrt{4\beta_k^2+1}}{2}$\\

  \hspace{0.2cm}  2.3 $Z_{k+1}= W_k + \frac{\beta_k-1}{\beta_{k+1}}(W_k-W_{k-1})$\\
3.    $ k\leftarrow k+1 $ \\

\textbf{Until} stopping criterion is satisfied\\
4.    $ W_l^{t+1}= W_k $\\ \hline
\end{tabular*}
\end{table}

Previous studies demonstrated that NeNMF can overcome the problem of low convergence rate and numerical instability \cite{Guan2012}. Luckily, SDNMF/L shares the common necessary properties required by Nesterov's optimal algorithm as NeNMF. In other words, the objective function $F(W_l,H_l)$ is convex with respect to $H_l$ or $W_l$ respectively, and the corresponding gradient is Lipschitz continous. Thus, we adopted Nesterov's accelerated gradient descent algorithm with the following parameters related to $W_l$ and $H_l$ to solve \eqref{sub_l}:
\begin{eqnarray}\label{L_GradH}
  \nabla_H F(W_l^t,H) = -W_l^t H_{l-1}+ W_l^tW_lH,
\end{eqnarray}
and
\begin{eqnarray}\label{L_GradW}
  \nabla_W F(W,H_l^t) = - H_{l-1}H_l^t+ WH_lH_l^t+\mu _l \xi_l^t\xi_lW.
\end{eqnarray}
We illustrated the pre-training procedure (\textbf{Algorithm 1} in TABLE \ref{INI_SDNMF/L}), which alternatively minimizes one matrix factor with another one fixed until convergence. Particularly, in the $l$-th layer, suppose there have been $t$ pairs of $W_l$ and $H_l$, to get the next iteration point $H_l^{t+1}$, \textbf{Subalgorithm 1} solves a sub-problem by constructing two sub-sequences like that in problem (\ref{NeNMF_H approximal fun}) and updates them until convergence. $W_l^{t+1}$ is obtained similarly with \textbf{Subalgorithm 2}. This process proceeds alternatively until convergence.

After pre-training each layer separately, the algorithm fine-tunes the system with initial points of all $W_l$, $H_l$ to reduce the total reconstruction error as follows:
\begin{eqnarray}\label{SDNMF/L}
                     & \min & C_{\mbox{SDNMF/L}}= \frac{1}{2}\parallel X-W_1W_2...W_LH_L\parallel^2_F \notag \\
                     &      &+\frac{1}{2}\Sigma^L_{l=1}\mu_l \mbox{tr}((\xi_l W_l)^T(\xi_l W_l)) \\
                                                                                               \notag \\
                     & \mbox{s.t.} & H_l\approx W_{l+1}H_{l+1}, \quad \forall \quad l=1,2,...,L-1,  \notag\\
                     &      &W_l \geq 0, H_l \geq 0, \quad \forall \quad l=1,2,...,L \notag.
\end{eqnarray}

\noindent
\textbf{Update rule for $W$ during the fine-tuning process} \\
For $W_l$, the subproblem in fine-tune process is equivalent to the following one,
\begin{eqnarray}\label{L_W_l}
                     &\min  &\frac{1}{2}\parallel X-W_1W_2...W_{l-1}W\widetilde{H_l} \parallel_F^2 \notag \\
                     &      &+ \frac{1}{2}\lambda_l \mbox{tr}((\xi_l\widetilde{W_l})^T(\xi_l\widetilde{W_l}))  \\
                                                                                                   \notag \\
                     & \mbox{s.t.} & W  \notag \geq 0,
\end{eqnarray}
where $\widetilde{H_l}$ is the reconstruction of $H_l$ and $\widetilde{H_l}\approx \widetilde{W_{l+1}}\widetilde{H_{l+1}}$. The objective function is convex with respect to $W$, and the gradient
\begin{equation}\label{SDNMF/L_gradW}
   \nabla_W F(\Psi, W,\widetilde{H}_l)= -\Psi_{l-1}^T X\widetilde{H}_l^T+\Psi_{l-1}^T \Psi_{l-1}W\widetilde{H}_l\widetilde{H}_l^T+\mu_l\xi_l^T\xi_l W, \\
\end{equation}
is Lipschitz continuous with Lipschitz constant $LC_{W_l}=\parallel \Psi_{l-1}^T\Psi_{l-1}\parallel_2 \cdot \parallel \widetilde{H_l} \widetilde{H_l}^T\parallel_2+ \mu_l \parallel \xi_l^T \xi_l\parallel_2$, where $\Psi_{l-1} = W_1W_2...W_{l-1}$. Thus, this problem can be solved by \textbf{Subalgorithm 2}, where the $F(W,H_l^t)$ should be substituted with the one in \eqref{L_W_l}, and $H_l$, $ LC_{W_l}$ should be substituted with $\widetilde{H_l}$ and $ LC_{W_l}=\parallel \Psi_{l-1}^T\Psi_{l-1}\parallel_2 \cdot \parallel \widetilde{H_l} \widetilde{H_l}^T\parallel_2+ \mu_l \parallel \xi_l^T \xi_l\parallel_2$ respectively.

\noindent
\textbf{Update rule for $H$ during the fine-tuning process}
For $H_l$, the subproblem is formulated as follows,
\begin{eqnarray}\label{L_H_l}
                     &\min & \frac{1}{2}\parallel X-\Psi_l H \parallel_F^2 \notag\\
                     & \mbox{s.t.} & H \geq 0. \notag
\end{eqnarray}
The objective function above is convex with respect to $H$ and its gradient
 \begin{equation}\label{SDNMF/L_gradH}
  \nabla_H F(\Psi_l,H)= -\Psi_l ^T X + \Psi_l^T \Psi_l H,
\end{equation}
 is Lipschitz continuous with the constant $LC_{H_l} = \parallel \Psi_l^T\Psi_l\parallel_2$, where $\Psi_l= \Psi_{l-1}W_l$. We utilized \textbf{Subalgorithm 1} to update $H_l$, where the objective function $ F(W_l^t,H)$ should be substituted with the one in \eqref{L_H_l}, the $W_l^t$ and the Lipschitz constant should be substituted with $\Psi_l$ and $LC_{H_l} = \parallel \Psi_l^T\Psi_l\parallel_2$ respectively. Finally, we proposed an algorithm for SDNMF/L in TABLE \ref{Finetune_SDNMF/L} based on the above analysis.

\begin{table}[htb] \small
\caption{}\label{Finetune_SDNMF/L}
\centering  
\begin{tabular*}{250pt}{lccc}  
\hline
 Algorithm 2: Optimal algorithm for SDNMF/L \\ \hline
\textbf{Input}:         $X \in R^{m\times n}$, the size of each layer \\
\textbf{Output}:        Weight matrices $W_l$ and representation matrices $H_l$ ($\forall$ $l$) \\
\textbf{Initialize all layers}:\\
      \hspace{0.5cm}      for all layers do\\
      \hspace{0.8cm}         $ (W_l,H_l)= \mbox{Algorithm1}(H_{l-1},\mbox{layers}(l)) $\\
      \hspace{0.5cm}      end for\\
\textbf{end initialization
}  \\
\textbf{Repeat}\\
      \hspace{0.5cm}            for all layers do\\
      \hspace{0.8cm}             \begin{numcases}{\widetilde{H_l}=}
                                   H_l, & for $l = L$ \notag\\
                                  W_{l+1}\widetilde{H_{l+1}}, & for $l< L$. \notag
                                 \end{numcases} \\
      \hspace{0.8cm}                        $\Psi_{l-1}=W_1W_2...W_{l-1} $ \\
      \hspace{0.8cm}                        $ W_l   \leftarrow \mbox{Subalgorithm2} ( W_l,\widetilde{H_l}, \Psi_{l-1} , \nabla_WF(W,\widetilde{H_l}), LC_{W_l}) $\\

      \hspace{0.8cm}                        $ \Psi_l\leftarrow \Psi_{l-1} W_l $ \\
      \hspace{0.8cm}                        $ H_l   \leftarrow \mbox{Subalgorithm1} (H_l, \Psi_l,\nabla_HF(\Psi_l,H), LC_{H_l}) $ \\
      \hspace{0.5cm}           end for \\
\textbf{Until} stopping criterion is satisfied.\\
\hline
\end{tabular*}
\end{table}

\noindent
\textbf{Algorithm complexity analysis}\\
During the initialization process, to update $H_l$, we need to compute $LC_{H_l^t}$ in \textbf{subalgorithm 1}. Without loss of generality, we let $l=1$. The complexity of computing $LC_{H_1^t}$ is $O(mk_1^2+k_1^3)$, where $m$ is the number of rows of data $X$; $k_1$ denotes the number of sub-basis in the first layer. The complexity of computing $H_k$ in \textbf{subalgorithm 1} is $O(mnk_1)+I_{inner}\cdot(nk_1^2)$, where $I_{inner}$ denotes the number of iterations in \textbf{subalgorithm 1}. Thus, the complexity to get $H_1^{t+1}$ through \textbf{subalgorithm 1} is $O(mk_1^2+k_1^3+mnk_1)+I_{inner}\cdot(nk_1^2)$. Similarly, the complexity to get $W_1^{t+1}$ is $O(nk_1^2+k_1^3+mnk_1)+I_{inner}\cdot(mk_1^2)$ in \textbf{subalgorithm 2}, where the complexity of computing $LC_{W_1^t}$ is dominated by computing $\parallel(H_1^t)(H_1^t)^T\parallel_2$ because $\xi_1^T\xi_1$ is one matrix with all unit elements whose $\parallel\cdot\parallel_2$ is the length of $\xi_1$. Thus, the complexity of initialization in the first layer is $I_{out}\cdot(O(mk_1^2+nk_1^2+mnk_1)+I_{inner}\cdot O(mk_1^2+nk_1^2))$, where $I_{out}$ is the number of iterations in \textbf{Algorithm 1}. Let $R=\max \{k_1,k_2,...,k_L\}$, the complexity of initialization for each layers is $I_{out}(O(mR^2+nR^2+mnR)+I_{inner}\cdot O(mR^2+nR^2))$.

During the fine-tuning process in \textbf{Algorithm 2}, we updated each factor while keeping the others fixed and the process began from $W_1$. Similar to the initialization, the complexity of tuning the two factors in each layer is $O(mR^2+nR^2+mnR)+f_{inner}\cdot(mR^2+nR^2)$, where $f_{inner}$ is the iteration number in corresponding sub-algorithm to yield each factor.

In a word, the computing complexity of SDNMF/L is $L\cdot(I_{out}\cdot(O(mR^2+nR^2+mnR)+I_{inner}\cdot O(mR^2+nR^2)))+f_{out}\cdot(L\cdot(O(mR^2+nR^2+mnR)+f_{inner}\cdot(mR^2+nR^2)))$. Empirically, both $I_{inner}$ and $f_{inner}$ are very small. If we ignore the inner iteration in both processes, the computing complexity is the same as Deep Semi-NMF \cite{Trigeorgis2017}.

\noindent
\textbf{Convergence analysis}\\
According to \cite{Grippo2000}, nonlinear Gauss-Seidel method under convex constraints will converge to the critical point (where the gradient is zero) eventually if two conditions are satisfied. For two-block coordinate problem, if the objective function of each subproblem is convex and the sequence generated by the algorithm has limit points, then every limit point is a critical point of the objective function. However, in a $m$-block ($m\geq3$) coordinate problem, the conditions must be that the objective function of $f(x_1,x_2,...,x_m)$ is strictly quasi-convex with respect to $m-2$ variables (or blocks), and the sequence generated by the algorithm has limit points.

In our framework, when $L\geq 3$, we have $m\geq4$. To analyze the convergence of SDNMF/L model, we consider the existence of the limit points of the sequence generated by SDNMF/L first. The fact that the objective is decreasing under the sequence supports that they are in a level set of $C_{\mbox{SDNMF/L}}$. $C_{\mbox{SDNMF/L}}$ is continuous, so this level set is closed. If this level set is unbounded, then $C_{\mbox{SDNMF/L}}$ is unbounded on this level set, contradicting with the definition of level set. Thus, the level set is a bounded and closed set (compact set). Correspondingly, the generated sequence within a compact set has limit points. Although we cannot demonstrate the final convergence since the strict quasi-convexity of $C_{\mbox{SDNMF/L}}$ is hard to prove, it does decrease after each iteration, and eventually converges to a local optima.

\subsubsection{SDNMF/R}
Here, we considered to control the $L_1$-norm of columns of each $H_l$ to deal with a sparse coding problem. It helps to represent the raw data with as fewer bases as possible while maintaining the ability to discriminate samples of different classes. Specifically, SDNMF/R is formulated as follows,
\begin{eqnarray}\label{SDNMF/R}
                     &\min   &C_{\mbox{SDNMF/R}}=  \frac{1}{2}\parallel X - W_1W_2...W_LH_L\parallel_F^2 \notag \\
                     &       & + \frac{1}{2}\Sigma_{l=1}^{L}\lambda_l \mbox{tr}((e_lH_l)^T(e_lH_l))\\
                                                                                                  \notag \\
                     & \mbox{s.t.} & H_l\approx W_{l+1}H_{l+1},   \notag \\
                     &    &W_l \geq0, H_l\geq 0, \forall l=1,2,...,L-1. \notag
\end{eqnarray}
Similarly, we adopted an initialization procedure like SDNMF/L using \textbf{Algorithm 1} to accelerate the optimization, and then we fine-tuned all $W_l$ and $H_l$ to reduce the total reconstruction error. During the fine-tuning process, for a specific $l$, the following two subproblems were considered to finetune $W_l$ and $H_l$:
\begin{eqnarray}\label{R_W_l}
                 W_l=& \arg \min_{W\geq 0} \frac{1}{2}\parallel\ X-\Psi_{l-1} W \widetilde{H}_l\parallel_F^2, \notag \\
\end{eqnarray}
\begin{eqnarray}\label{R_H_l}
               H_l=  & \arg\min_{H\geq0}\frac{1}{2} \parallel X- \Psi_l H  \parallel_F^2 \notag \\
                     & + \lambda_l \mbox{tr}(e_lH_l)^T(e_lH_l),
\end{eqnarray}
where $\Psi_{l-1}=W_1W_2...W_{l-1}$ and $\Psi_l=\Psi_{l-1}W_l$.

Subproblem \eqref{R_W_l} can be solved by \textbf{Subalgorithm 2}. The objective function, $H$ and Lipschitz constant should be substituted with the one in \eqref{R_W_l}, $\widetilde{H_l}$ and $LC_{W_l}=\parallel\Psi_{l-1}^T\Psi_{l-1}\parallel_2\cdot\parallel\widetilde{H}_l\widetilde{H}_l^T\parallel_2$, respectively. Subproblem \eqref{R_H_l} can be solved by \textbf{Subalgorithm 1} by substituting the objective function, $W$ and Lipschitz constant with the one in \eqref{R_H_l}, $\Psi_l=\Psi_{l-1}W_l$ and $L_{H_l}=\parallel \Psi_l^T\Psi_l\parallel_2+ \lambda_l \parallel e_l^Te_l\parallel_2$, respectively. Thus, similar to SDNMF/L, with corresponding parameters in \textbf{Algorithm 2} replaced, SDNMF/R will find its solution.

\subsubsection{SDNMF/RL1}
We next considered to control the sparsity of both $W$ and $H$. It generates sparse basis matrices as well as a sparse representation.
\begin{eqnarray}\label{SDNMF/RL1}
                     &\min   C_{\mbox{SDNMF/RL1}}= & \frac{1}{2}\parallel X - W_1W_2...W_LH_L\parallel_F^2 \notag \\
                     &                      & + \frac{1}{2}\Sigma_{l=1}^{L}\mu_l \mbox{tr}((\xi_lW_l)^T(\xi_lW_l)) \notag \\
                     &                      & + \frac{1}{2}\lambda_L \mbox{tr}((e_LH_L)^T(e_LH_L))\\
                                                                                                                      \notag \\
                     & \mbox{s.t.}  & H_l\approx W_{l+1}H_{l+1},   \notag \\
                     &              & W_l \geq0, H_l\geq 0, \forall l=1,2,...,L-1. \notag
\end{eqnarray}
\noindent
To solve \eqref{SDNMF/RL1}, initialization for each layer is also necessary. For a specific $l$, the problem is:
\begin{eqnarray}\label{SDNMF/RL1_INI}
                     & \min &\frac{1}{2}\parallel H_{l-1}-W_l H_l\parallel^2_F + \frac{1}{2}\mu_l \mbox{tr}((\xi_lW_l)^T(\xi_lW_l)) \notag \\
                     &      & +\frac{1}{2}\lambda_l \mbox{tr}((e_lH_l)^T(e_lH_l)) \\
                                                                                                                               \notag \\
                     & \mbox{s.t.} &W_l \geq 0, H_l \geq 0. \notag
\end{eqnarray}
where $\frac{1}{2}\lambda_l \mbox{tr}((e_lH_l)^T(e_lH_l))$ is considered only when $l=L$. Obviously, the objective function of \eqref{SDNMF/RL1_INI} is convex with respect to $W$ or $H$ separately and the respective Lipschitz constant is easy to calculate. Given the fact, back to \textbf{Algorithm 1}, the solution for \eqref{SDNMF/RL1_INI} is ready to obtain with parameters related to $W$ and $H$ substituted with the following ones:
\begin{equation}\label{SDNMF/RL1_INI_W}
W:  \begin{cases}
                           \nabla_W F(W,H_l) &= -H_{l-1}H_l^T+WH_lH_l^T+\mu_l \xi_l^T\xi_lW,                         \\
                             LC_{W}           &=  \parallel H_lH_l^T\parallel_2+\mu_l \parallel \xi_l^T \xi_l\parallel_2,
\end{cases}
\end{equation}
\begin{equation}\label{SDNMF/RL1_INI_H}
H:  \begin{cases}
                           \nabla_H F(W_l,H) &= -W_l^TH_{l-1}+W_l^TW_lH+\lambda_l e_l^Te_lH,                         \\
                             LC_{H}           &=  \parallel W_l^TW_l\parallel_2+\lambda_l \parallel e_l^T e_l\parallel_2,
\end{cases}
\end{equation}
where $\lambda_l e_l^Te_lH $ and $\lambda_l\cdot\parallel e_l^T e_l\parallel_2$ in \eqref{SDNMF/RL1_INI_H} are removed if $l\neq L$, meaning that we just controled the sparsity of $H_L$ to obtain a high-level final presentation. After pre-training each layer separately, we fine-tuned the weights of all layers as well as the final representation with initial approximation points of $W_l$, $H_L$ to reduce the total reconstruction error of \eqref{SDNMF/RL1}.
In particular, for a specific $l$ and a factor matrix $W_l$, let's consider the model in \eqref{L_W_l} where,
\begin{equation}\label{SDNMF/RL1_TUNE_W}
W:  \begin{cases}
                           \nabla_W F(W,\widetilde{H}_l) &= -\Psi_{l-1}^TX\widetilde{H}_l^T+\Psi_{l-1}^T\Psi_{l-1} W \widetilde{H}_l\widetilde{H}_l^T \\
                                                         &  +\mu_l \xi_l^T\xi_lW,  \\
                             LC_{W}                       &=  \parallel \widetilde{H}_l\widetilde{H}_l^T\parallel_2+\mu_l \parallel \xi_l^T \xi_l\parallel_2,
\end{cases}
\end{equation}
By replacing the corresponding parameters in \textbf{Subalgorithm 2} with equation \eqref{SDNMF/RL1_TUNE_W}, a new point for $W_l$ in the fine-tuning process will be obtained.
For $H_L$, consider the model \eqref{R_H_l}, where
\begin{equation}\label{SDNMF/RL1_TUNE_H}
H:  \begin{cases}
                           \nabla_H F(\Psi_l,H) &= -\Psi_L^TX+\Psi_L^T\Psi_LH+\lambda_L e_L^Te_LH,                          \\
                             LC_{H}              &=  \parallel \Psi_L^T\Psi_L\parallel_2+\lambda_L \parallel e_L^T e_L\parallel_2.
\end{cases}
\end{equation}
Similarly, with the above given parameters, a new point for $H_L$ can also be obtained using \textbf{Subalgorithm 1}, indicating that SDNMF/RL1 can also be optimized with \textbf{Algorithm 2}.

\subsubsection{SDNMF/RL2}
The above SDNMF/RL1 is devoted to find sparse basis as well as a sparse representation of raw data. Intuitively, for a decomposition of $X \approx WH$, if $W$ is compelled to be sparse, $H$ will tend to be smooth. Here, we tried to strengthen this trend by controlling the $L_1$-norm of each column of all weights while exerting $L_2$-norm constraints onto final $H$ (denoted as SDNMF/RL2).
\begin{eqnarray}\label{SDNMF/RL2}
                     &\min   C_{\mbox{SDNMF/RL2}}= &\frac{1}{2}\parallel X - W_1W_2...W_L H_L\parallel_F^2 \notag \\
                     &                      &+ \frac{1}{2}\Sigma_{l=1}^{L}\mu_l \mbox{tr}((\xi_l W_l)^T(\xi_l W_l)) \notag \\
                     &                      & + \frac{1}{2}\lambda_L \parallel H_L \parallel_F^2\\
                                                                                                             \notag \\
                     & \mbox{s.t.} & H_l\approx W_{l+1}H_{l+1}, \quad W_l \geq0,   \notag \\
                     &         &    H_L\geq 0, \forall l=1,2,...,L-1. \notag
\end{eqnarray}
The optimization procedure, either initialization or fine-tuning, is similar to SDNMF/RL1 as long as the $e_L^Te_L$ in \eqref{SDNMF/RL1_INI_H} and \eqref{SDNMF/RL1_TUNE_H} are replaced by identity matrix $I$ in the optimization of \eqref{SDNMF/RL2}.

\subsubsection{Nonlinear function}
In this part, we took $g(x)$ to be a nonlinear function and considered four choices: tanh, root, sigmoid, softplus. Due to the incorporation of nonlinear functions, it is different from linear system to solve nonlinear system, but the process was still divided into initialization and the fine-tuning steps. In the initialization, at first, we decomposed $X=W_1H_1$, before using $H_1$ into the next layer, we projected it in an element-wise way with the nonlinear function. Similarly, all $H_l$ $(l=2,3,...,L-1)$ will be nonlinearly projected before making it as the input of the next layer. As for $H_L$, we can project it or just leave it unchanged, resulting in two ways of incorporation. After projection, the strategy to solve $H_{l-1}=W_lH_l$ is the same as that in Table \ref{INI_SDNMF/L}.

However, with nonlinear structure, Lipschitz constant for the gradient of nonlinear functions is hard to obtain, which means that Nesterov's optimal method may be not suitable to fine-tune the system. Thus, we fine-tuned the system with a traditional project gradient descend algorithm. The gradient for each factor is computed as follows:
\begin{eqnarray}\label{gradient_H_L}
                     &\frac{\partial f}{\partial H_L}= &W_L^T\frac{\partial f}{W_LH_L}\notag \\
                     &                               = &W_L^T\left( \frac{\partial f}{g^{-1}(W_LH_L)}\odot \nabla g^{-1}(W_LH_L)\right)\notag \\
                     &                               = & W_L^T\left(\frac{\partial f}{H_{L-1}}\odot \nabla g^{-1}(W_LH_L)\right),
\end{eqnarray}
\noindent
\begin{eqnarray}\label{gradient_W_l}
                     &\frac{\partial f}{\partial W_l}= &\frac{\partial f}{W_lH_l}H_l^T\notag \\
                     &                               = &\left( \frac{\partial f}{g^{-1}(W_lH_l)}\odot \nabla g^{-1}(W_lH_l)\right)H_l^T\notag \\
                     &                               = &\left(\frac{\partial f}{H_{l-1}}\odot \nabla g^{-1}(W_lH_l)\right)H_l^T.
\end{eqnarray}
We updated $H_L$ first with (29), which can be computed according to the chain rule. Then, we updated $W_l,l\in \{2,...,L\}$ with (30). As for $H_l, l\in\{1,2,...,L-1\}$, it is approximated directly by the nonlinear projection of the product of $W_{l+1}$ and $H_{l+1}$, i.e., $H_l\approx g^{-1}(W_{l+1}H_{l+1})$. For $W_1$, we updated it with \textbf{Subalgorithm 2} in Table \ref{INI_SDNMF/L}.

\section{Experiment}
\subsection{Data}
\noindent
We applied our models (DNMF, SDNMF/R, SDNMF/L, SDNMF/RL1, SDNMF/RL2) onto two benchmarking image datasets: ORL and PIE-pose 27.0, and compared them with other NMF-related models. Specifically, we compared our models with single layer models: the projected gradient NMF (PgNMF), NeNMF, and deep matrix factorization models: Deep Semi-NMF and Multi-layer models. Generally, a large number of samples are required to train a deep model to learn the complex features. Otherwise, it may not be able to outperform single layer models. We tested our models firstly on data ORL. It contains 40 subjects and each subject owns 10 images of equal size $112\times92$ under different conditions. Thus, the data size of ORL is $10304\times 400$. It was randomly split into a training set with size of $10304\times 320$ and a test set with size of $10304\times 80$. The second dataset is PIE-pose 27.0. It contains 68 subjects and each subject owns 42 images of equal size of $32\times32$ under each condition. Thus, the size of this data is $1024\times 2856$. In our experiments, PIE-pose 27.0 was randomly split into seven independent pairs of training and test sets for cross validation, where the size of each training set was $1024\times 2448$ and the size for test set was $1024\times 408$. With these two datasets, we first investigated the structure of our models to figure out the proper number of layers as well as the number of sub-bases in the hidden layers.

\subsection{Evaluation criteria}
\noindent
In this article, we adopted three measures including normalized mutual information (NMI) \cite{Danon2005}, error rate(ER) \cite{Lin2009} and naive precision (NP). Given the standard class partition $C^*$ and the obtained class partition $C$, we first constructed a confusion matrix $N$ whose rows correspond to the classes in $C^*$ and columns correspond to the classes in $C$. Let $N_{ij}$ denotes the number of samples overlapped by the $i$-th real and $j$-th obtained classes. NMI is defined as
\begin{eqnarray}\label{NMI}
                     &\mbox{NMI}(C,C^*)=\frac{-2\Sigma_{i=1}^{|C|}\Sigma_{j=1}^{|C^*|}N_{ij}log(\frac{N_{ij}N}{N_{i.}N_{.j}})}{\Sigma_{i=1}^{|C|}N_{i.}log(\frac{N_{i.}}{N})+\Sigma_{j=1}^{|C^*|}N_{.j}log(\frac{N_{.j}}{N})},\notag
\end{eqnarray}
where $|c|$ is the number of classes in $C$; $N_i.$ is the sum of $i$-th row of $N$; $N_.j$ is the sum of $j$-th column of $N$.
ER is defined as follows,
\begin{eqnarray}\label{errorate}
                     & \mbox{ER}(Z,Z^*)=\sqrt{\parallel Z^*(Z^*)'-ZZ' \parallel}, \notag
\end{eqnarray}
where the $Z$ and $Z^*$ are the indicator matrix for $C$ and $C^*$.
NP is computed as follows:
\begin{eqnarray}\label{NaivePrecison}
                     & \mbox{NP}(C,C^*)=\frac{1}{|C^*|}\Sigma^{|C^*|}_{i=1}\frac{\max_{j\in\{1,2,...,|C^*|\}}N_{ij}}{|c^*_i|}, \notag
\end{eqnarray}
where $c^*_i$ denotes the number of objects in $i$-th classes of $C^*$.

\subsection{Results of ORL data}
\begin{figure}[!htbp]
  \includegraphics[scale=0.39]{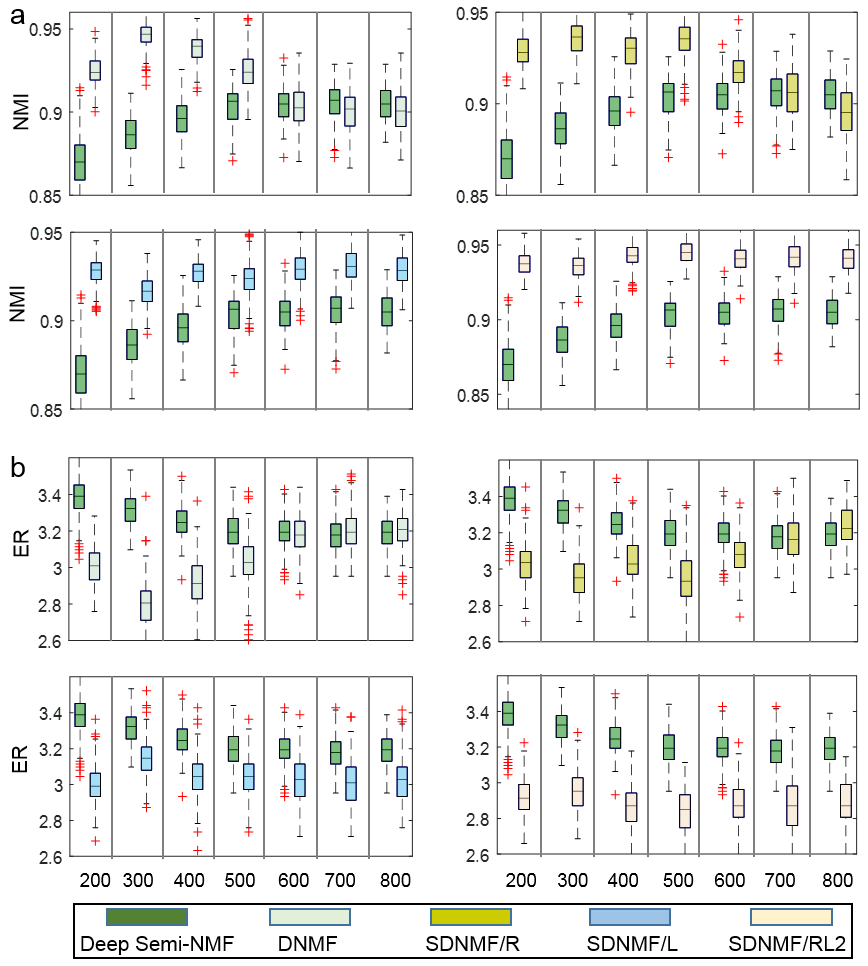}
  \caption{Comparison of two-layer models with layersize=$[k_1, 160$] in terms of NMI (a) and ER (b). Each sub-figure shows the results of Deep Semi-NMF and our models for comparison.}\label{fig1}
\end{figure}

\begin{figure}[!htbp]
 \centering
  \includegraphics[scale=0.45]{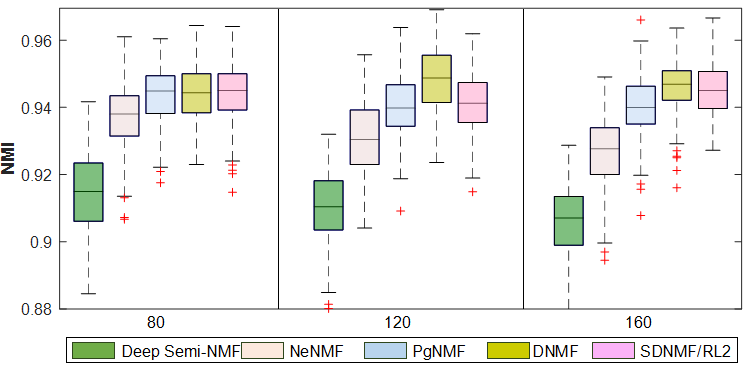}
  \caption{Comparison of PgNMF, NeNMF, Deep Semi-NMF, DNMF and SDNMF/RL2 in terms of NMI under different $k_2$. Under each $k_2$, we chose the best $k_1$ among 100,200,300,400,500,600,700,800 for each deep model.}\label{fig2}
\end{figure}

\noindent
The ORL data has relative small number of samples, which is insufficient to train a deep model. Thus, we initially set $L=2$. We chose $k_2=80,120,160$ and $k_1=100,200,300,400,500,600,700,800$. In order to see the role of an extra layer, we fix $k_2$ while making $k_1$ change, that is, for each $k_2$ we ran our models with $k_1$ varying from 100 to 800. Figure \ref{fig1} shows the NMI and ER of all models under $k_2=160$, from which we select appropriate value of $k_1$ for all models. For example, we chose $k_1=300$ for DNMF and SDNMF/R, $k_1=700$ for SDNMF/L, $k_1=500$ for SDNMF/RL2. Similarly, we chose appropriate $k_1$ for $k_2$=80,120 for each model including Deep Semi-NMF. For Deep Semi-NMF, we fixed $k_1=700$ according to its performance across all tested values. After selection, we compared our sparse deep models with Deep Semi-NMF and single layer models (Figure \ref{fig2}). Each of boxes in both Figure \ref{fig1} and \ref{fig2} represents the distribution of 200 precision scores. They were generated by 20 computing experiments on ORL and $K$-means was applied 10 times on the representation matrix of each experiment for robust classification. Results show that the classification ability of our models outperforms Deep Semi-NMF, but do not always perform better than PgNMF and NeNMF. These results indeed confirm our consideration that ORL data does not contains enough samples to train a deep model sufficiently.

\subsection{Results of PIE data}
\subsubsection{Structure optimization}
In previous work, the number of layers is specified without a thorough consideration. Different dataset and algorithms may require different structures. Similar to that in Deep Semi-NMF \cite{Trigeorgis2017}, we first conducted a lot of experiments on the whole PIE-pose 27.0 dataset (without splitting the images into training and testing sets) to compare the classification performance of all models under different number of layers ($L=2,3,4$). For each model, each value of $L$ is tested by 50 experiments. The number of sub-bases in hidden layer in each experiment differs from one to another. They are randomly chosen from exponential distributions and then ranked to make sure that the value in lower layer is larger than that in higher layer. To make a roughly fair comparison, the number of sub-bases in the last layer is fixed to be 40.
\begin{figure}[!htbp]
  \centering
  \includegraphics[scale=0.34]{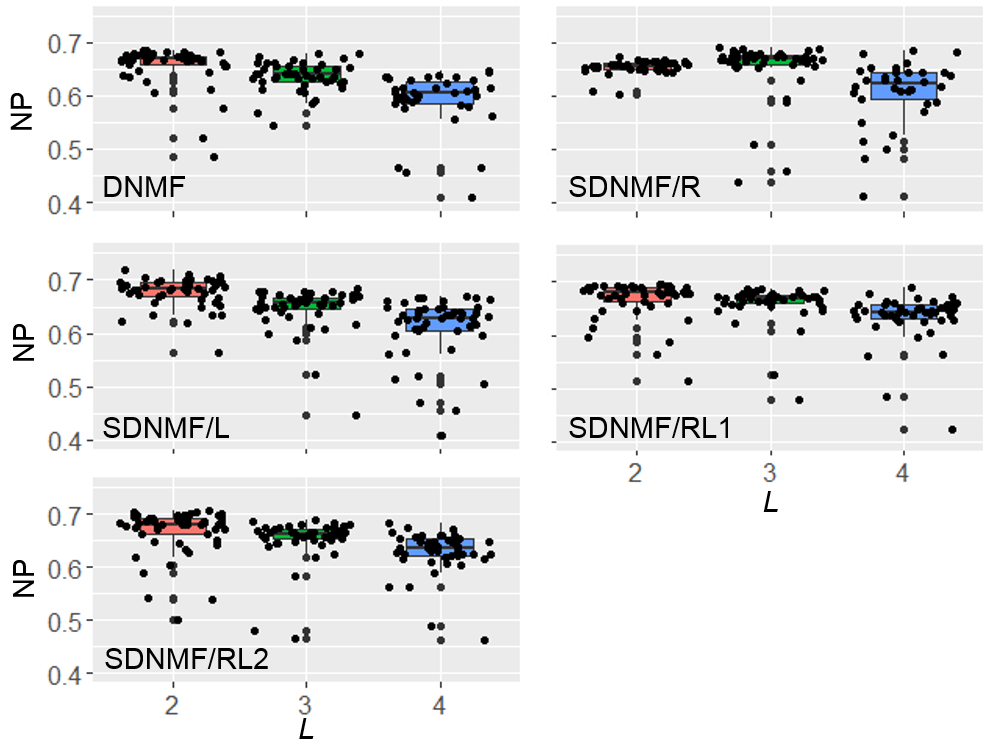}
  \caption{Classification precision of our linear models on PIE-pose 27.0 dataset in terms of NP with different layer number $L$= 2, 3, 4. The number of sub-bases in hidden layers was randomly extracted from certain exponential distributions. Then they were ranked to satisfy the relationship $k_{lower layer}>k_{higher layer}$. For example, for SDNMF/L with $L=3$ and $layersize=[k_1,k_2,k_3]$, the value of $k_1$, $k_2$ were drawn from exponential distributions, with $k_1>k_2$, $k_3$ is fixed to be 40. The experiments for SDNMF/L under each layer size was repeated 50 times, generating 50 NMI values. The rest experiments for the other models were done in the same way.}\label{fig3}
\end{figure}

\begin{figure}[!htbp]
  \centering
  \includegraphics[scale=0.49]{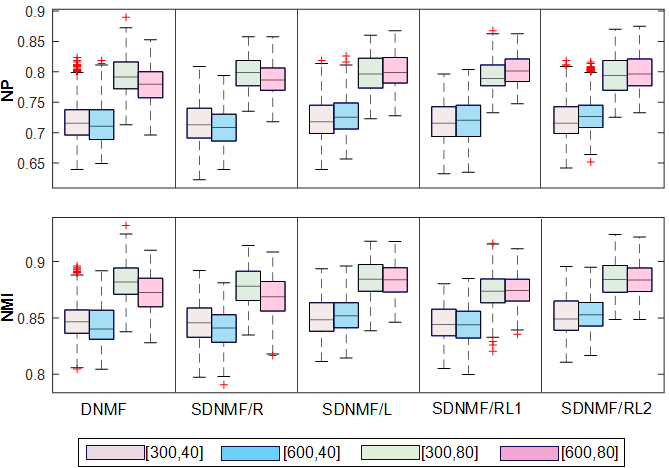}
  \caption{$k_1$ selection of our proposed linear deep models on PIE-pose 27.0 dataset. Each linear model was tested under four conditions: layersize=$[300,40],[300,80],[600,40],[600,80]$.}\label{fig4}
\end{figure}

\begin{figure}[!htbp]
  \centering
  \includegraphics[scale=0.45]{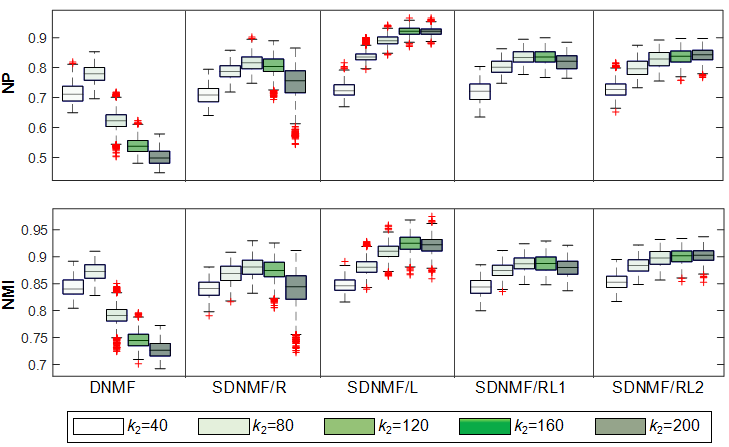}
  \caption{$k_2$ selection of our proposed linear deep models on PIE-pose 27.0 dataset. Each linear model was tested under five conditions: layersize=$[600,40],[600,80],[600,120],[600,160],[600,200]$.}\label{fig5}
\end{figure}

\begin{figure}[!htbp]
  \centering
  \includegraphics[scale=0.49]{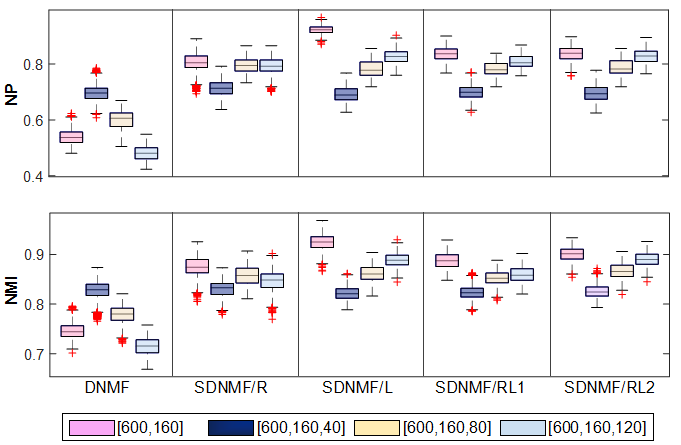}
  \caption{Correspondent comparison between two-layer and three layer linear deep models on PIE-pose 27.0 dataset. Specifically, the layer size for three-layer model contains $[600,160,40]$, $[600,160,80]$ and $[600,160,120]$. The layer size for our two-layer models is $[600,160]$.}\label{fig6}
\end{figure}

From Figure \ref{fig3}, we can see that $L=2$ should be the best choice for PIE-pose 27.0 data classification. To figure out the specific structure for our models, we made a series of detailed experiments on the randomly split 7 independent pairs of training and testing datasets of PIE. We compared the results of each model with layer size $[300,40],[300,80],[600,40],[600,80]$ to determine the value of $k_1$ (Figure \ref{fig4}). All of the boxes related to the PIE pose-27.0 in the following experiments (Figure \ref{fig4}, \ref{fig5}, \ref{fig6} and \ref{fig7}) represent the distribution of 1400 precision scores: Under each condition, for each pair of training and testing datasets, each model runs 20 times and $K$-means is adopted 10 times for each running to seek robust classification performance.

Generally, the classification results with $k_1=300$ and $k_1=600$ are consistent (Figure \ref{fig4}). To strengthen the ability of learning necessary information in the first layer, we chose $k_1=600$. As for $k_2$, we tested it from $40,80,120,160,200$ (Figure \ref{fig5}).
It shows that, besides DNMF (It is one case of our models where there is no sparsity constraint on any $W_l$ or $H_l$), the precision of four models rise as $k_2$ grows at first and reach the climax under $k_2=120$ or $k_2=160$ and then fall down. Let $k_2=160$, an extra layer is added to the existing models and we compared three-layer models with two-layer ones (Figure \ref{fig6}).
It shows that, most of our models, except DNMF, reach their best with layer size$=[600,160]$. Given that DNMF does not generate better results against our other sparse deep models, we did not extend all models to three-layer level for PIE-pose 27.0 data.

\subsubsection{Role of nonlinear function}
\begin{figure}[!htbp]
  \includegraphics[scale=0.48]{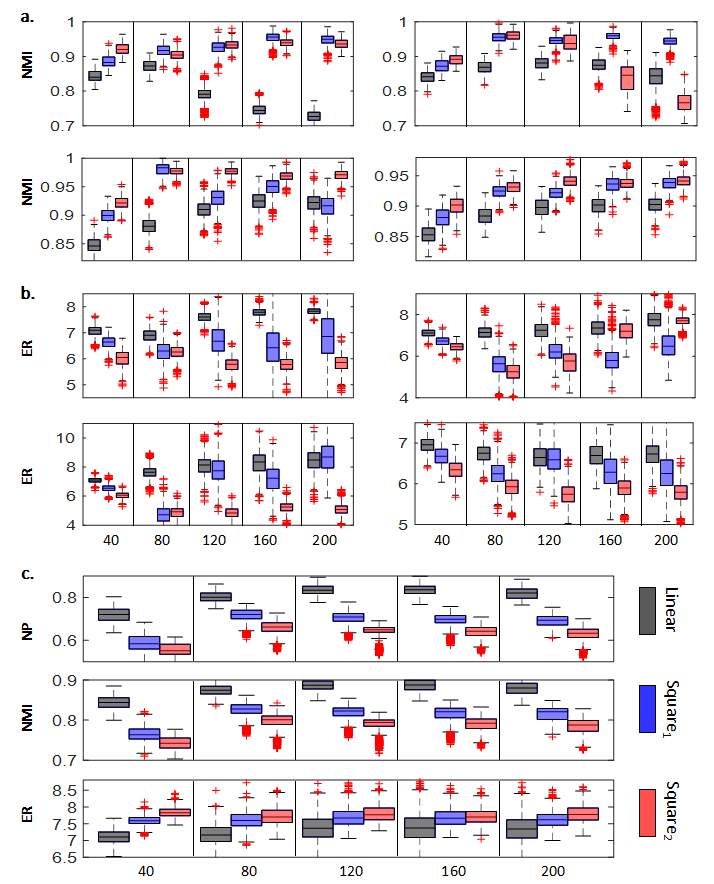}
  \caption{Comparison among linear and nonlinear models. $\mbox{Square}_1$ means that we only added nonlinear function onto $H_1$, whereas $\mbox{Square}_2$ means that both $H_1$ and $H_2$ are projected by the nonlinear function. The classification precision in terms of NMI of DNMF, SDNMF/R, SDNMF/L, SDNMF/RL2 are shown in (a) and (b), respectively. The results of SDNMF/RL1 in terms of NP, NMI, ER are shown in (c).}\label{fig7}
\end{figure}

In this section, we incorporated nonlinear functions (tanh, root, sigmoid, softplus) into our models to achieve more powerful representation ability. After inspection, only root function was left. We took $g(x)=\sqrt{x}$ and incorporated it into the original linear models with $L=2$ and layersize=[600,160]. Adding nonlinear functions means that, before using $H$ as the input of the next layer or as the final usage for classification, we projected it nonlinearly first. Since $L=2$, we tried two ways to incorporate the root function into our models. One was that we only projected $H_1$, leaving $H_2$ unchanged. The other way was that we projected both $H_1$ and $H_2$. The steps to solve the nonlinear models are similar to those for linear models. We first initialized each layer and fine-tuned the whole system. We compared the performance of each nonlinear model with corresponding linear models (Figure \ref{fig7}).

It shows that there are significant improvements of DNMF, SDNMF/L, SDNMF/RL2 in terms of NMI and ER after inducing $g(x)=\sqrt{x}$. Specifically, for DNMF, the changing trend of NMI and ER of its linear model are even reversed by nonlinear function. As for SDNMF/R, its learning ability is strengthened by projecting $H_1$ but damaged by projecting both $H_1$ and $H_2$. Maybe it is inappropriate to project $H_2$ with $g(x)=\sqrt{x}$ since, according to the design and corresponding results of our experiments, the elements of $H_2$ are smaller than 1 so that the projection of $g(x)=\sqrt{x}$ will enlarge all the elements and make $H_2$ more smooth, undermining the original sparsity generated by the our initial sparse constriction. For SDNMF/RL1, we required sparsity on both $W$ and $H$. This may cause the loss of important information and cannot realize a good approximation. This situation might be deteriorated by projecting $H_1$ or $H_1$ and $H_2$ with the root function. We compared our models with PgNMF, NeNMF, Deep Semi-NMF and Multi-layer NMF ($\theta=0.001$) in terms of NMI, NP and ER (Table \ref{ComparisonP}).

\begin{table}[htb]
\centering
\scriptsize
\begin{threeparttable}
\caption{Comparison of NMF-related models on PIE-pose 27 data} \label{ComparisonP}
\begin{tabular*}{240pt}{@{\extracolsep{\fill}}ccccccccc}
  \toprule
  Method                  &NMI &NP &ER   \\
  \midrule
  PgNMF                & 0.884  &  0.786   & 6.691       \\
  NeNMF                & 0.914  &  0.868   &6.729        \\
  Deep Semi-NMF        & 0.945 &  0.910    &5.772        \\
 \\
  Multi-layer NMF (linear)             & 0.905    &0.845   &6.616        \\
  Multi-layer NMF ($\mbox{Square}_1$)  & 0.909    & 0.851  &6.332        \\
  Multi-layer NMF ($\mbox{Square}_2$)  & 0.888    & 0.813  &6.510       \\
 \\
  SDNMF/L (linear)             & 0.925  &0.922    &8.271        \\
  SDNMF/L ($\mbox{Square}_1$)  & 0.950  & 0.946   &7.195         \\
  SDNMF/L ($\mbox{Square}_2$)  & \textbf{0.968 } & \textbf{0.951}   &\textbf{5.234}      \\
  \\
  SDNMF/RL2 (linear)             & 0.901  &0.837   &6.706        \\
  SDNMF/RL2 ($\mbox{Square}_1$)  & 0.936  & 0.910  &6.320        \\
  SDNMF/RL2 ($\mbox{Square}_2$)  & 0.937  & 0.902  &5.882       \\
  \\
  SDNMF/R (linear)             & 0.875  &0.805   &7.353        \\
  SDNMF/R ($\mbox{Square}_1$)  & 0.958  & 0.942  &5.857        \\
  SDNMF/R ($\mbox{Square}_2$)  & 0.837  & 0.722  &7.20       \\
  \\
  DNMF (linear)             & 0.750  &0.543   &7.786        \\
  DNMF ($\mbox{Square}_1$)  & 0.954  & 0.943  &6.485        \\
  DNMF ($\mbox{Square}_2$)  & 0.940  & 0.903  &5.779       \\
  \\
  SDNMF/RL1 (linear)             & 0.890  &0.836   &7.390        \\
  SDNMF/RL1 ($\mbox{Square}_1$)  & 0.814  & 0.690  &7.674        \\
  SDNMF/RL1 ($\mbox{Square}_2$)  & 0.789  & 0.636  &7.717       \\
  \bottomrule
\end{tabular*}
\begin{tablenotes}
        \footnotesize
        \item[1] All results from deep models were generated under layersize=[600,160] where as results of PgNMF and NeNMF were obtained with $k$=160.
\end{tablenotes}
\end{threeparttable}
\end{table}
It turns out that SDNMF/L outperforms PgNMF, NeNMF and Multi-layer NMF in terms of NP and NMI. Advantages against the single-layer NMF mean that the deep decompositions of SDNMF/L works better to analyze data whereas advantages against Multi-layer NMF mean that both of the sparse strategy and the optimization strategies of SDNMF/L play important roles in generating a higher level data representation for more accurate classification. Sparsity constraints for all the columns of $W$ helps to extract localized features and learn class-specific deep features, which in turn generates more distinctive representations for classification. Also, the fine-tuning process in the optimization approximates original data more closely. What's more, the performance of Multi-layer NMF varies greatly with respect to parameter $\Theta$. We tested it under multiple choices and some of the results were shown in Appendix. As for the comparison with Deep Semi-NMF, there is a gap between linear SDNMF/L and Deep Semi-NMF. This is reasonable since the learning ability is limited by the nonnegativity of our models whereas Semi-NMF is more likely to succeed without that constriction on basis matrices. Nevertheless, by adopting root function, the learning ability of SDNMF/L rises to a new level, generating more representative coefficient matrices than Deep Semi-NMF to distinguish samples of different classes. The ER of SDNMF/L is higher than those of the compared models. We see that sometimes multiple (up to 5) classes would be clustered into the same one by linear SDNMF/L, which rarely occurs when incorporating the root function (especially in the second way). Moreover, the learning ability of nonlinear SDNMF/RL2, SDNMF/R as well as DNMF is also greatly strengthened by such a nonlinear function.

\subsubsection{Feature hierarchies}
\begin{figure*}[!htbp]
 \centering
  \includegraphics[scale=0.51]{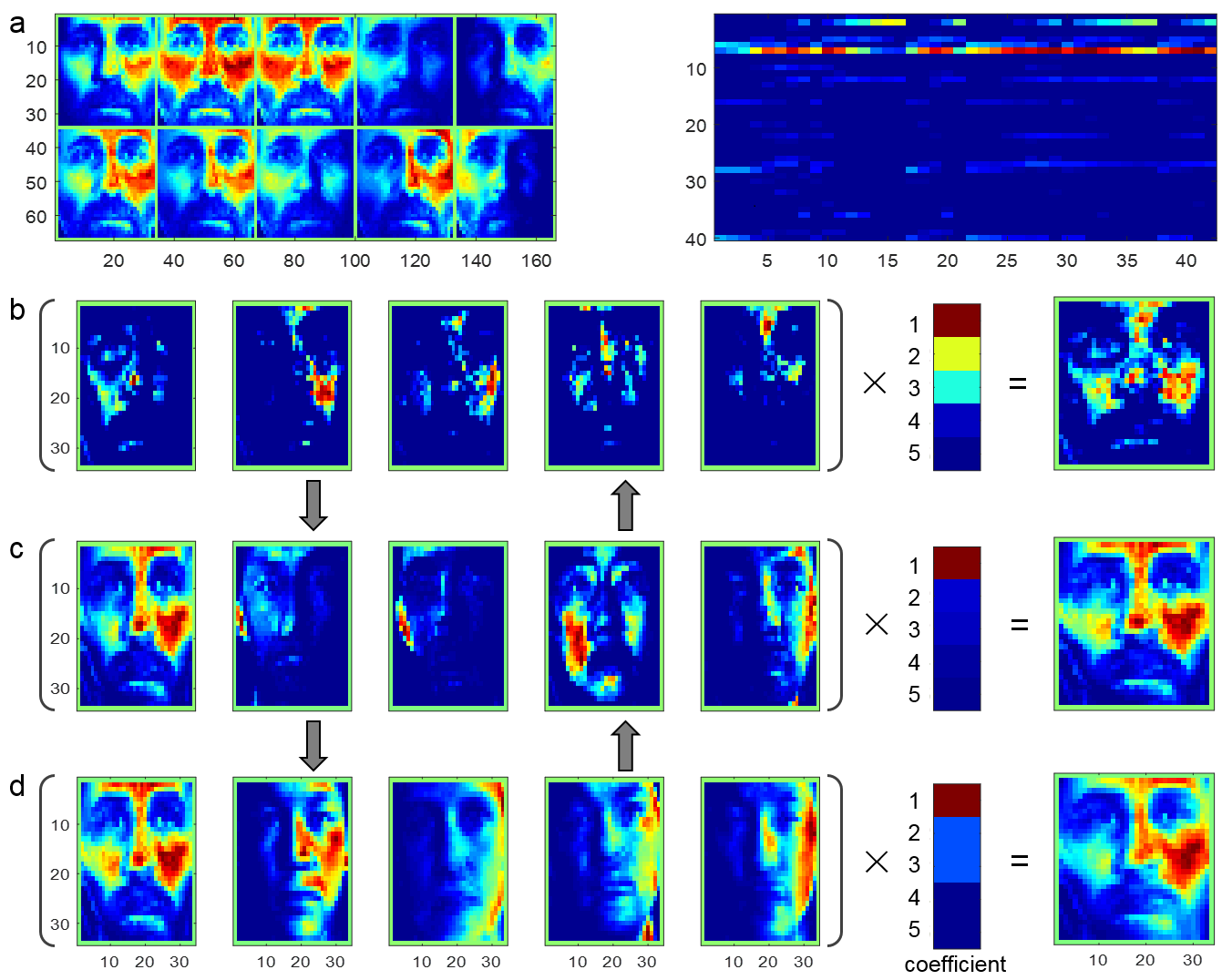}
  \caption{Hierarchical feature interpretation by a linear SDNMF/L model with layersize=[193,141,40]. (a) 10 samples of class 36 and the representation matrix of class 36. Certain bases as well as corresponding coefficients in the first, second and third layer are shown in (b), (c) and (d), respectively.}\label{fig8}
\end{figure*}

SDNMF/L can not only yield a good classification on data PIE-pose 27.0, but also learns feature hierarchies to show parts-of-whole characteristics intuitively. This benefits from our sparsity constraints on the columns of $W$. Sparse $W_1$ helps to learn part-based information for each sample and sparse $W_2$ selectively combine the initial basis in the first layer to generate composite basis and form relatively complex features. Again, $W_3$ selectively combine the composite basis in the second layer to show discriminative features for samples in distinct classes. As all the features being extracted, a high level and more meaningful representations for each class will be automatically obtained to realize accurate classification and feature interpretation. Figure \ref{fig8} shows how a three-layer linear SDNMF/L model learns feature hierarchies. In Figure \ref{fig8}a, the left part contains 10 samples of class 36 and the right part are coefficients (some columns of $H_3$) for all samples of class 36, from which we knew that samples of class 36 are mainly reconstructed by the 7th composite basis in $W_1W_2W_3$ (the first face image in Figure \ref{fig8}d). Since it is the combination result of the composite bases in $W_1W_2$ in the second layer, we got the top 5 elements of the 7th column of $W_3$ (coefficient in Figure \ref{fig8}c) and located the corresponding composite basis in $W_1W_2$ (the 5 face images in the left in Figure \ref{fig8}c). To find the related initial sub-basis in the first layer, we chose the first image in Figure \ref{fig8}c, which has the largest coefficient and located its column number in $W_1W_2$ (column 13). Then we ranked the 13th column of $W_2$ and get the top 5 elements (coefficient in Figure \ref{fig8}b) and located the corresponding columns of the $W_1$ (the 5 images in the left in Figure \ref{fig8}b).

\section{Conclusion}
\noindent
In this paper, we proposed sparse deep nonnegative matrix factorization models satisfying different sparsity requirements. By extending the original NMF into multi-layer models, they can learn data in a hierarchical way. Model structure optimization was implemented for different datasets. We explored the incorporation of nonlinear functions into these models. We adopted the Nesterov's accelerated algorithm to accelerate the computing process during optimization. We evaluated the time complexity of our algorithm framework and showed that it is comparable to Deep Semi-NMF. We demonstrated that our models are able to learn more discriminative representations to realize competitive classification accuracy compared with other NMF models. Besides, they can also generate intuitive interpretations for the features extracted across all layers, which is not applicable for single layer NMF or Deep Semi-NMF.

We would like to close this paper by listing some possible research directions. Our models performed differently even on the same dataset, there must be some underlying reasons that need to be explored. Another concern is why some nonlinear functions performs well while others not. Besides, although the complexity of our model are comparable to other NMF variants, we need to search for more efficient optimization strategies to deal with the increasing big data.

\section*{Acknowledgment}
This work has been supported by the National Natural Science Foundation of China [No. 61422309, 61379092, 61621003 and 11661141019]; the Strategic Priority Research Program of the Chinese Academy of Sciences (CAS) [XDB13040600], CAS Frontier Science Research Key Project for Top Young Scientist [No. QYZDB-SSW-SYS008], the Outstanding Young Scientist Program of CAS, and the Key Laboratory of Random Complex Structures and Data Science, CAS [No. 2008DP173182].



%


\end{document}